\documentclass[letterpaper, 10 pt, conference]{ieeeconf}  

\IEEEoverridecommandlockouts                              

\usepackage{graphics} 
\usepackage{epsfig} 
\usepackage{mathptmx} 
\usepackage{times} 
\usepackage{amsmath} 
\usepackage{amssymb}  
\usepackage{pifont}
\usepackage{bm}
\usepackage{multirow}
\usepackage{graphicx}
\usepackage[final]{changes}
\usepackage{optidef}
\usepackage{array}
\usepackage{colortbl}
\DeclareGraphicsExtensions{.pdf,png}
\title{\LARGE \bf
\added{PPF: Pre-training and Preservative Fine-tuning of Humanoid Locomotion via Model-Assumption-based Regularization}
}

\author{Hyunyoung Jung$^{*,1}$, Zhaoyuan Gu$^{*,1}$, Ye Zhao$^{1}$, Hae-Won Park$^{2}$ and Sehoon Ha$^{1}$
\thanks{*These two authors contribute equally to work.}
\thanks{$^{1}$Georgia Institute of Technology, Atlanta,
        GA, 30308, USA
        {\tt\small {hjung331@gatech.edu, zgu78@gatech.edu, ye.zhao@me.gatech.edu, sehoonha@gatech.edu}}}%
\thanks{$^{2}$Korea Advanced Institute of Science and Technology, Yuseong-gu,
        Daejeon, 34141, Republic of Korea
        {\tt\small {haewonpark}@kaist.ac.kr}}%
\thanks{This paper has been accepted for publication in IEEE Robotics and Automation Letters.}
}
\begin{document}
\bstctlcite{IEEEexample:BSTcontrol}

\maketitle
\thispagestyle{empty}
\pagestyle{empty}



\newcommand{\cmt}[1]{}

\newcommand{\updated}[1]{\textcolor{blue}{{#1}}}
\newcommand{\hyunyoung}[1]{\textcolor{orange}{{Hyunyoung: #1}}}
\newcommand{\donghoon}[1]{\textcolor{cyan}{{Zhaoyuan: #1}}} 
\newcommand{\sehoon}[1]{\textcolor{red}{{Sehoon: #1}}}

\newcommand{\newtext}[1]{#1}
\newcommand{\eqnref}[1]{Equation~(\ref{eq:#1})}
\newcommand{\figref}[1]{Figure~\ref{fig:#1}}
\newcommand{\tabref}[1]{Table~\ref{tab:#1}}
\newcommand{\secref}[1]{Section~\ref{sec:#1}}

\long\def\ignorethis#1{}

\newcommand{\etal}{{\em{et~al.}\ }}
\newcommand{\eg}{e.g.\ }
\newcommand{\ie}{i.e.\ }

\newcommand{\figtodo}[1]{\framebox[0.8\columnwidth]{\rule{0pt}{1in}#1}}



\newcommand{\pdd}[3]{\ensuremath{\frac{\partial^2{#1}}{\partial{#2}\,\partial{#3}}}}

\newcommand{\mat}[1]{\ensuremath{\mathbf{#1}}}
\newcommand{\set}[1]{\ensuremath{\mathcal{#1}}}

\newcommand{\vc}[1]{\ensuremath{\mathbf{#1}}}
\newcommand{\vEndEff}{\ensuremath{\vc{d}}}
\newcommand{\vRelMove}{\ensuremath{\vc{r}}}
\newcommand{\sSet}{\ensuremath{S}}

\newcommand{\vControl}{\ensuremath{\vc{u}}}
\newcommand{\vPoint}{\ensuremath{\vc{p}}}
\newcommand{\sSpringCoef}{{\ensuremath{k_{s}}}}
\newcommand{\sDamperCoef}{{\ensuremath{k_{d}}}}
\newcommand{\vHandle}{\ensuremath{\vc{h}}}
\newcommand{\vForce}{\ensuremath{\vc{f}}}

\newcommand{\mTransChain}{\ensuremath{\vc{W}}}
\newcommand{\mRotateTrans}{\ensuremath{\vc{R}}}
\newcommand{\sJoint}{\ensuremath{q}}
\newcommand{\vJoint}{\ensuremath{\vc{q}}}
\newcommand{\mJoint}{\ensuremath{\vc{Q}}}
\newcommand{\mMass}{\ensuremath{\vc{M}}}
\newcommand{\sMass}{\ensuremath{{m}}}
\newcommand{\vGravity}{\ensuremath{\vc{g}}}
\newcommand{\vConstr}{\ensuremath{\vc{C}}}
\newcommand{\sConstr}{\ensuremath{C}}
\newcommand{\vCOM}{\ensuremath{\vc{x}}}
\newcommand{\sGeneralForce}[1]{\ensuremath{Q_{#1}}}
\newcommand{\vStateVar}{\ensuremath{\vc{y}}}
\newcommand{\vControlVar}{\ensuremath{\vc{u}}}
\newcommand{\tr}[1]{\ensuremath{\mathrm{tr}\left(#1\right)}}

%
%

\renewcommand{\choose}[2]{\ensuremath{\left(\begin{array}{c} #1 \\ #2 \end{array} \right )}}

\newcommand{\gauss}[3]{\ensuremath{\mathcal{N}(#1 | #2 ; #3)}}

\newcommand{\pctab}{\hspace{0.2in}}
\newenvironment{pseudocode} {\begin{center} \begin{minipage}{\textwidth}
                             \normalsize \vspace{-2\baselineskip} \begin{tabbing}
                             \pctab \= \pctab \= \pctab \= \pctab \=
                             \pctab \= \pctab \= \pctab \= \pctab \= \\}
                            {\end{tabbing} \vspace{-2\baselineskip}
                             \end{minipage} \end{center}}
\newenvironment{items}      {\begin{list}{$\bullet$}
                              {\setlength{\partopsep}{\parskip}
                                \setlength{\parsep}{\parskip}
                                \setlength{\topsep}{0pt}
                                \setlength{\itemsep}{0pt}
                                \settowidth{\labelwidth}{$\bullet$}
                                \setlength{\labelsep}{1ex}
                                \setlength{\leftmargin}{\labelwidth}
                                \addtolength{\leftmargin}{\labelsep}
                                }
                              }
                            {\end{list}}
\newcommand{\newfun}[3]{\noindent\vspace{0pt}\fbox{\begin{minipage}{3.3truein}\vspace{#1}~ {#3}~\vspace{12pt}\end{minipage}}\vspace{#2}}

\newcommand{\key}{\textbf}
\newcommand{\fun}{\textsc}


%
%

\newcommand{\ieeeEq}[1]{Eq.~(\ref{#1})} 
\newcommand{\eqnum}[1]{\tag{#1}}        

\newcommand{\citeet}[2]{#1 et al.~\cite{#2}}
\begin{abstract}
Humanoid locomotion is a challenging task due to its inherent complexity and high-dimensional dynamics, as well as the need to adapt to diverse and unpredictable environments.
In this work, we introduce a novel learning framework for effectively training a humanoid locomotion policy that imitates the behavior of a model-based controller while extending its capabilities to handle more complex locomotion tasks, such as more challenging terrain and higher velocity commands.
Our framework consists of three key components: pre-training through imitation of the model-based controller, fine-tuning via reinforcement learning, and model-assumption-based regularization (MAR) during fine-tuning. In particular, MAR aligns the policy with actions from the model-based controller only in states where the model assumption holds to prevent catastrophic forgetting.
We evaluate the proposed framework through comprehensive simulation tests and hardware experiments on a full-size humanoid robot, Digit, demonstrating a forward speed of 1.5 m/s and robust locomotion across diverse terrains, including slippery, sloped, uneven, and sandy terrains.
\end{abstract}

\section{Introduction}

Humanoid robots hold unique potential to operate seamlessly in human-centric environments. To realize this, they are expected to function reliably across a wide range of indoor and outdoor settings, which requires advanced locomotion capabilities. However, achieving robust locomotion in unstructured environments remains challenging due to hybrid dynamics involving complex contact and the high dimensionality of bipedal systems.
Previous works have approached this problem using model-based methods that compute foot placement using simplified models~\cite{Kajita2001,pratt2006capture}. To enhance constraint handling and stability, these simplified models are often integrated with optimization-based methods~\cite{li2023dynamic, gibson2022terrain, Scianca_TRO, gu2024walkingbylogic} in model predictive control (MPC). However, such approaches with simplified models can be less adaptable in complex environments. An alternative approach uses more accurate full-order models but at the cost of computation speed~\cite{Dai_centroidal_kinematics} or accuracy~\cite{khazoom2024tailoring}. In addition, the hybrid nature of bipedal locomotion dynamics complicates the optimization formulation, which is often restricted to fixed contact sequences.

More recently, bipedal locomotion has been tackled by learning-based approaches that train control policies through gradient descent on data-driven objective functions~\cite{Ilija2024realworld, li2023Robust, siekmann2021simtoreal, jkonah2021blind}. As demonstrated in quadrupedal locomotion~\cite{miki2022learning, lee2020learning, rudin2022learning, kumar2021rma, peng2020learning}, these learning-based approaches can enhance robustness and agility. However, they often require extensive reward engineering, and their lack of interpretability and long training times make interactive tuning difficult.
\begin{figure}[t]
\centering
\label{fig:hw_testbed}
\includegraphics[width=\columnwidth]{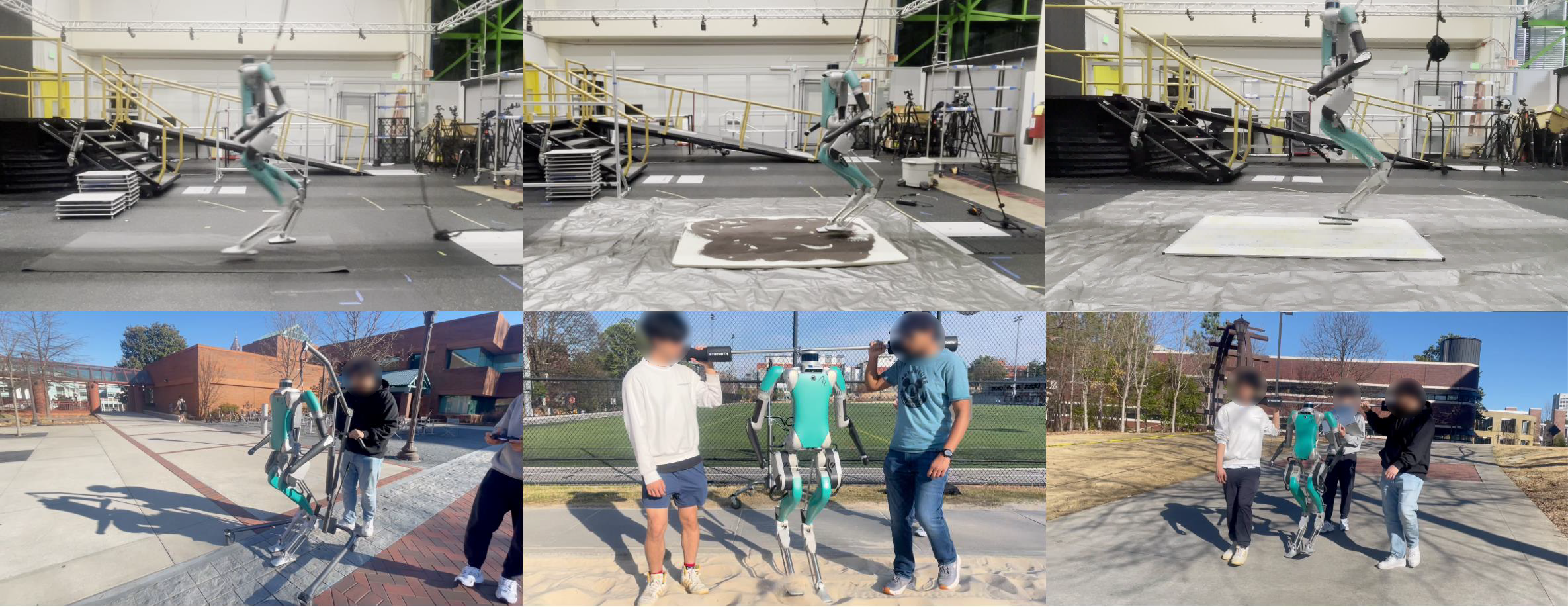}
\vspace{-2em}
\caption{Our \replaced{Pre-training and Preservative Fine-tuning (PPF)}{Pre-training and Continual Improvement (PreCi)} framework achieves a forward velocity of 1.5 m/s while successfully traversing a whiteboard covered with poppy seeds\added{ or olive oil}, as well as diverse outdoor terrains, including hills, uneven surfaces, and sand.}
\vspace{-2em}
\end{figure}
To address this, approaches that combine model-based and learning-based approaches have also been widely explored~\cite{xie2023glide, yang2022fast, da2021learning}. This hybrid strategy has also been shown to be effective for humanoids, particularly in applications involving contact planning~\cite{yu2024learninggenericdynamiclocomotion, lee2024integrating}. However, they often require a model-based controller to run in the backend, which can be expensive at runtime. A similar challenge was encountered earlier in the quadrupedal context and was tackled by entirely substituting the model-based controller with a behavior-cloned neural network policy~\cite{youm2023imitating, miller2023reinforcementlearningleggedrobots}. Especially, Youm et al.~\cite{youm2023imitating} demonstrated that a robust control policy can be efficiently trained using a two-stage learning process, where the first stage involves imitation of a model-based controller (MBC), followed by reinforcement learning (RL) to fine-tune performance.

Motivated by this two-stage learning process, we seek to train a robust humanoid locomotion policy that achieves higher speed and more robust locomotion while employing symmetric and periodic gaits of MBC.
However, unlike quadrupeds, humanoid has more unstable dynamics where its center of mass can easily shift out of its supporting polygon. When RL is applied for fine-tuning, the policy often suffers from catastrophic forgetting~\cite{wang2024acomprehensive}, overfitting to the task or dynamics it is trained on and deviating significantly from the motions learned during pre-training. This leads to a loss of the originally imitated behavior and degraded performance in the target domain.

To mitigate this problem, we introduce a novel learning framework, \replaced{PPF: Pre-training and Preservative Fine-tuning of a humanoid control policy via model-assumption-based regularization.}{PreCi: Pre-training and Continual Improvement of a humanoid control policy via model-assumption-based regularization.} Unlike the previous two-stage learning approach~\cite{youm2023imitating}, our method introduces an adaptive regularization term based on the model assumption violation in the fine-tuning stage. More specifically, our method regularizes the policy to match the model-based controller's actions if the robot’s state aligns with the assumptions of the underlying model. This approach is especially useful when the fine-tuning task extends beyond the capabilities of the model-based controller\deleted{, as it reduces regularization in states where the robot violates model assumptions}.\added{ For example, the ALIP model~\cite{gong2022zerodynamicspendulummodels} assumes a constant body height and zero vertical velocity. When these assumptions are violated, such as during high-speed motion or traversal over rough terrain, the regularization weight is reduced, allowing the policy to improve beyond the limitations of the model.}

We demonstrate \replaced{PPF}{PreCi} on a full-sized humanoid robot, Digit, in both simulation and real-world environments. In the simulation, our method can traverse both uneven and sloped terrains while achieving the highest tracking performance compared to the baseline methods.
On hardware, our method can achieve diverse locomotion tasks including fast forward walking and robust walking across various indoor and outdoor terrains. Specifically, \replaced{PPF}{PreCi} achieves a forward walking speed of 1.5 m/s while maintaining robustness on a whiteboard covered with poppy seeds\added{ or olive oil}, as well as on outdoor sloped, uneven, and deformable sandy terrains.
\section{Related Work}
\subsection{Model-based Control for Humanoid Locomotion}
Humanoid locomotion often relies on optimization algorithms that use mathematical models to capture the robot's essential dynamics. One of the most widely used models for bipedal locomotion is the linear inverted pendulum model (LIPM)~\cite{Kajita2001}, which provides analytical solutions for the foot placement to achieve a desired CoM trajectory~\cite{pratt2006capture, gu2024walkingbylogic}. However, this simplified model fails to capture detailed joint-level behaviors, often resulting in conservative or infeasible full-body motions. To address these limitations, researchers have proposed more accurate models considering the inertia of the robot, such as single rigid body models ~\cite{li2023dynamic} and centroidal dynamics ~\cite{orin2013centroidal, CD_MPC_locomotion}. These inertial-informed models enable the online planning of the contact location, force, and centroidal states and have achieved agile locomotion skills such as running~\cite{wensing_thesis} and jumping~\cite{li2024continuousdynamicbipedaljumping}. Recently, with advancements in computational power, it has become feasible to plan with kino-dynamics~\cite{Dai_centroidal_kinematics} or even full-body dynamics model~\cite{khazoom2024tailoring, WB_MPC_Dantec}. Despite their impressive performance, model-based locomotion methods typically require accurate robot dynamics, well-informed environment setup, and a detailed specification of the gait sequence\added{ to obtain this performance}, which causes significant manual effort. 

\subsection{Learning-based Approach for Humanoid Locomotion}
In recent years, there have been significant advancements in legged locomotion due to the emergence of deep reinforcement learning algorithms and the power of massively parallel simulation environments.
These learning frameworks initially demonstrated impressive performance in quadrupedal locomotion by optimizing policies based on carefully designed rewards~\cite{miki2022learning, lee2020learning, rudin2022learning}, imitating reference motions~\cite{peng2020learning}, or incorporating model-based controllers~\cite{xie2023glide,yang2022fast,da2021learning}. \added{In particular, incorporating model-based methods enables learning-based methods to acquire periodic and symmetric gaits in a more interpretable way, mitigating the tedious reward shaping.}
\replaced{The similar success on quadruped has been replicated on humanoid locomotion via learning-based frameworks.}{Following this success, researchers began actively investigating learning-based frameworks, presenting inherent instability and complexity.} Some \replaced{studies proposed}{research groups proposed} novel architectures~\cite{Ilija2024realworld, li2023Robust} to derive actions from contextual information embedded in observation histories. Others explored the use of demonstration data to streamline the training and mitigate the reward engineering bottleneck by leveraging model-based methods~\cite{miller2023mimoc, liu2024opt2skill} or state-only motion capture data~\cite{li2023Robust, he2024learning}.

\subsection{Domain Transfer for Learning-Based Legged Locomotion}
While model-based approaches typically design their control frameworks directly in the target domain (i.e., the real world) or within high-fidelity simulators~\cite{gong2021onestep,gong2019feedback}, learning-based approaches usually utilize the simulation data to facilitate the data generation process. However, this often struggles with the well-known sim-to-real gap, leading to overfitting to the training simulator. Therefore, transferring a trained policy to different domains remains a key challenge in learning-based methods. In the context of legged locomotion, this problem is handled in two different ways. The first approach is by utilizing target domain data~\cite{Kostrikov2023Demonstrating, smith2023learning, smith2022legged, ha2020learning, haarnoja2019learning}, where they directly train the policy in the target domain either through fine-tuning or end-to-end learning. However, this approach requires expensive data collection, especially for humanoids. The other way utilizes the scalability of simulations~\cite{Ilija2024realworld,siekmann2021simtoreal,Siekmann2021BlindBipedal}, where they extensively randomize the physical parameters during training to enable zero-shot deployment in the target domain. 
While these methods perform well in real-world scenarios, questions remain about their sample efficiency and the integration of existing controllers. In this work, we aim to address the domain transfer challenge by learning the periodic and cyclic gaits of the model-based controller without catastrophic forgetting.

\section{Background: Model-based Controller}
\label{sec:mbc}

Our framework begins with a given model-based controller (MBC), which is distilled into a learnable neural network and fine-tuned using reinforcement learning (RL). In our implementation, we adopt the humanoid locomotion controller in Shamsah et al~\cite{Shamsah2023Integrated}, which consists of a foot placement controller based on the angular-momentum linear inverted pendulum (ALIP) model~\cite{gong2022zerodynamicspendulummodels} and a passivity-based whole-body inverse dynamics controller~\cite{hamid2017passivitybasedcontrol}.

\noindent\textbf{Foot placement controller.}
The work of~\cite{Shamsah2023Integrated} captures the dynamics of a bipedal robot using the reduced-order ALIP model that consists of a center of mass~(CoM) and its connecting massless telescopic legs.

The ALIP model from \cite{gong2022zerodynamicspendulummodels} uses angular momentum about the stance foot as the contact point. Assuming constant CoM height $z$, the ALIP model has the following dynamics:

\begin{equation}\dot{\boldsymbol{x}}_c = \frac{\boldsymbol{L}}{mz}, \dot{\boldsymbol{L}} = mg\boldsymbol{x}_c + \boldsymbol{u}_a,
\end{equation}
where $\boldsymbol{x}_c$ is the horizontal CoM position in a frame attached to the contact point, $\boldsymbol{L}$ is angular momentum about the contact point, and $\boldsymbol{u}_a$ is the ankle torque. $m$ is the mass of the robot, and $g$ is the gravitational acceleration.
This model assumes the constant height $\bar{z}$, and zero velocity and zero acceleration in the vertical direction:
\begin{equation}
    \label{eq:model_ass}
    z=\bar{z}, \; \; \dot{z} = 0, \; \; \ddot{z} = 0,
\end{equation}
with $\bar{z} = 1.01$ m for Digit.
In this model, the desired foot placement with respect to the CoM position $\boldsymbol{x}_c$ can be computed by the one-step-ahead prediction:
\begin{equation}
    \boldsymbol{x}^{\text{des}} = \frac{\boldsymbol{L} \cosh(\omega T)/mz - \boldsymbol{v}_c^{\text{des}}}{\omega \sinh(\omega T)},
\end{equation}
where $T$ is the step duration of one walking step, $\omega$ is the natural frequency given by $\sqrt{g / \bar{z}}$, and $\boldsymbol{v}_c^{\text{des}}$ is the desired CoM velocity.

\noindent\textbf{Passivity-based controller.}
Given a desired foot placement, we solve full-body inverse kinematics to generate smooth joint trajectories $\boldsymbol{q}^{\text{des}}, \boldsymbol{\dot{q}}^{\text{des}}, \boldsymbol{\ddot{q}}^{\text{des}}$, where $\boldsymbol{q}$ represents the full joint state of the robot, including both unactuated floating-base joints and actuated motor joints.

The inverse dynamics controller solves linearized dynamics $M \boldsymbol{\ddot{q}}^{\text{des}} - J^T \boldsymbol{\lambda} - S^T \boldsymbol{\tau} = -C\boldsymbol{\dot{q}} - G$, where $M$ is mass matrix, $J$ is contact jacobian, $S$ is selection matrix, $C$ is Coriolis and centrifugal term, and $G$ is the gravity vector. $\boldsymbol{\tau}$ is the joint torque, and \added{$\boldsymbol{\lambda}$} is the contact force. 
The inverse dynamics controller separates the linearized dynamics into actuated and unactuated parts to eliminate the contact force and solve for the desired joint torque. A passivity-based feedback controller is applied for stabilization~\cite{hamid2017passivitybasedcontrol} as feedback.
 
\section{Background: Imitating and Finetuning Model-based Controller}
\label{sec:ifm}

Our framework follows the approach of pretraining the policy using a model-based controller and fine-tuning it using reinforcement learning, as proposed by Youm et al.~\cite{youm2023imitating}. 
\noindent\textbf{Imitation of the model-based controller.}
In the first stage, the expert model-based controller is distilled to the learnable neural network using behavior cloning with Dataset Aggregation (DAgger)~\cite{ross2011reduction}.
The DAgger loss is given by the squared Euclidean norm between the actions:
\begin{equation} \label{eq:imitation}
    L_{DAgger} = \sum_{\left( \boldsymbol{s}, \boldsymbol{a^E}\right) \in \mathcal{D}}{ \left\| \boldsymbol{a}^E - \added{\mu_{\theta}(\boldsymbol{s})} \right\| ^2},
\end{equation}
where $\mathcal{D}$ is the aggregated data buffer, $\boldsymbol{s}$ and \added{$\boldsymbol{a}^E$} \replaced{denote}{are} the \replaced{sampled robot state}{current robot state} and \replaced{the corresponding action from the expert model-based controller\deleted{, respectively}, where $ E$ stands for the ``expert''}{the action of the given model-based controller}. \replaced{$\mu_{\theta}$ denotes the neural network policy trained via supervised learning. While deterministic here, it later serves as the mean of the Gaussian policy during RL fine-tuning.}{$\mu$ is the mean network of the policy.} This imitation gives us a pre-trained policy $\pi$ that behaves similarly to the model-based controller but also is learnable.

\noindent\textbf{Finetuning of the pre-trained policy.}
After pre-training the policy, its performance can be further improved through RL methods, such as Proximal Policy Optimization~\cite{schulman2017proximalpolicyoptimizationalgorithms} for the given task reward. 
The implementation also includes a velocity curriculum in the forward direction similar to~\cite{margolisyang2022rapid} and a terrain curriculum over five different terrains with increasing difficulties, including uphill, downhill, flat terrain, and uneven terrains~\cite{Ilija2024realworld, rudin2022learning}.
Lastly, the framework randomizes the dynamic parameters and adds noises to observations for sim-to-real transfer.

\section{\added{PPF: Pre-training and Preservative Fine-tuning}}

In this section, \replaced{we introduce Pre-training and Preservative Fine-tuning (PPF), a novel framework designed to enhance the performance of a given model-based controller by first distilling its knowledge into a learnable neural network, and then fine-tuning it using RL with model-assumption-based regularization.}{we introduce a novel framework of Pre-training and Continual Improvement (PreCi), which is designed to improve the performance of the given model-based controller via pre-training and continual improvement.} While our work is based on \added{Imitating and Finetuning Model Predictive Control (IFM)}~\cite{youm2023imitating}, IFM often suffers from catastrophic forgetting, which results in poor performance in the real-world environment (Section~\ref{sec:forgetting}). To overcome this issue, we design a regularized loss that further distills knowledge from the expert controller during fine-tuning (Section~\ref{sec:regularization}). In addition, we introduce a model-assumption-based regularization (MAR) to determine the reliability of the expert controller (Section~\ref{sec:assumption-based}). This allows us to reject unreliable data from MBC and thus improves the performance of the fine-tuned policy. Finally, we will describe a few implementation details, such as reward functions or Lipschitz Continuity Penalty (Section~\ref{sec:details}). 

\subsection{Motivation: Catastrophic Forgetting of Motion Style}
\label{sec:forgetting}
IFM demonstrated robust performance improvements in quadrupedal locomotion. However, it often suffers from catastrophic forgetting during the fine-tuning phase due to unconstrained policy optimization. This forgetting of motion style is even more critical for bipedal locomotion tasks because of their unstable dynamics. 

\begin{figure}
\centering
\vspace{0.5em}
\begin{tabular}{cc}
    \includegraphics[width=0.25\columnwidth]{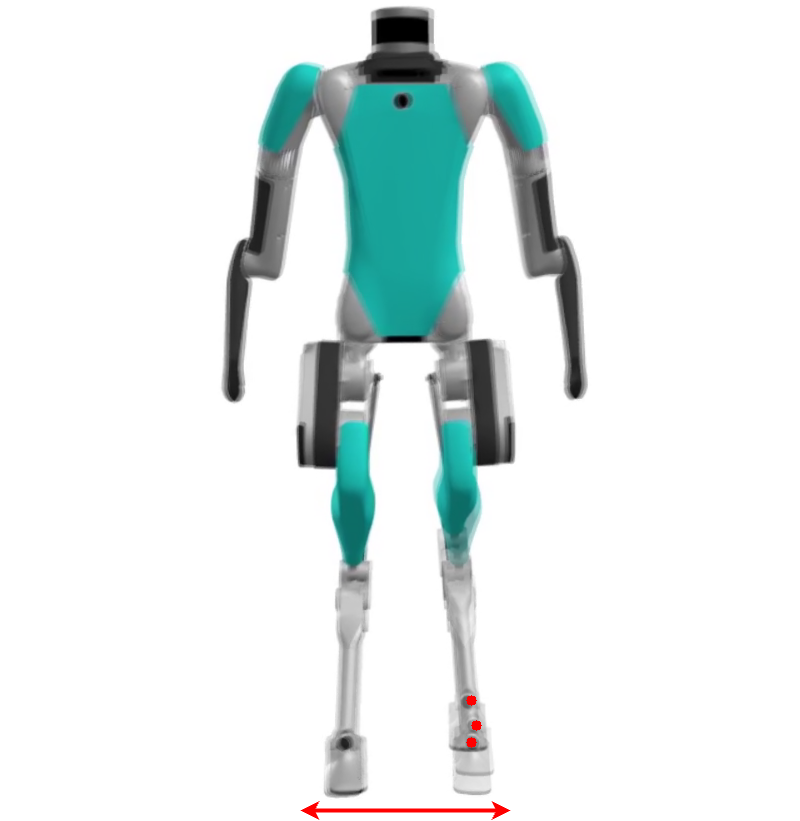} & 
    \includegraphics[width=0.25\columnwidth]{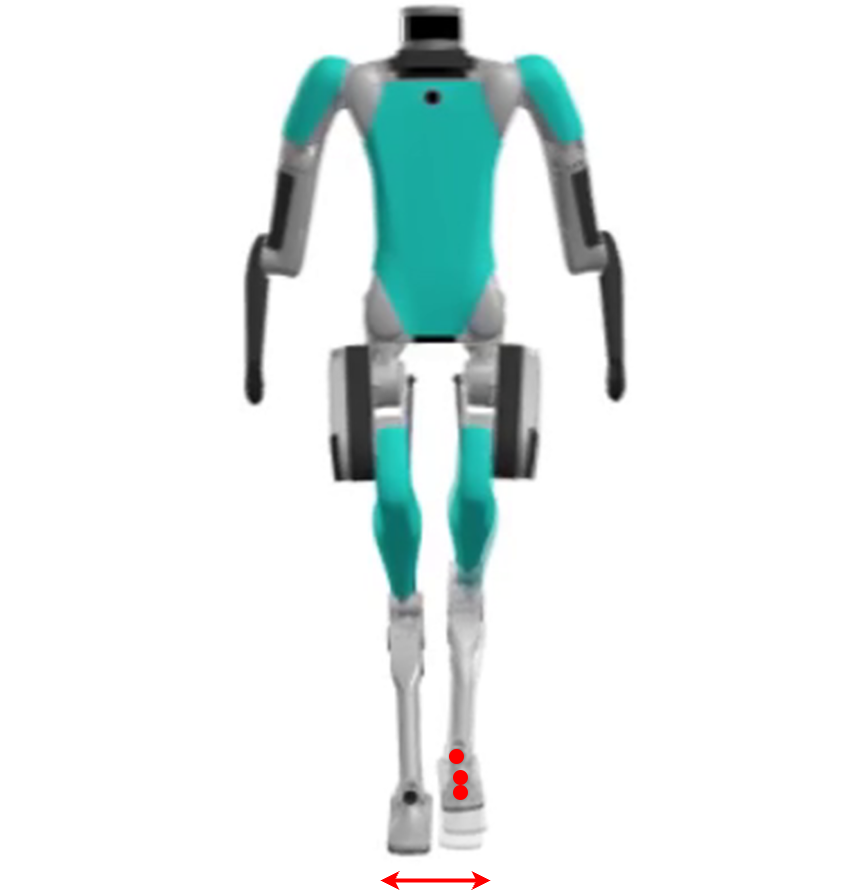} \\
    \scriptsize (a) MBC motion example & \scriptsize (b) IFM motion example \\
\end{tabular}
\vspace{-0.5em}
\caption{Motion forgetting example. Unlike MBC, IFM is trained \replaced{to swing its foot inwards initially and step with narrower foot placement, optimizing for lateral tracking accuracy and reduced energy consumption.}{to narrow its foot in the swing phase to optimize the tracking reward in the lateral direction.}}
\vspace{-2em}
\label{fig:forgetting_example}
\end{figure}

\replaced{During RL-based fine-tuning, IFM often stabilizes the pelvis lateral movement by initially swinging the leg inward and stepping with narrower foot placement as shown in Fig.~\ref{fig:forgetting_example}.}{During RL-based fine-tuning, IFM often stabilizes its torso movement by adjusting swing leg abduction inward and reducing lateral momentum as shown in Fig.~\ref{fig:forgetting_example}.} However, this overfitted adaptation in a training environment can lead to potential instability\deleted{ due to the reduced size of the contact area}, and this problem will be further exacerbated when the policy is deployed on hardware with a significant sim-to-real gap. \added{For instance, frequent foot collisions were observed when deploying the IFM in both the AR-sim and on hardware.} While domain randomization may help mitigate this issue, its performance improvement is marginal and requires extensive manual tuning through trial and error.

\subsection{Improved Fine-tuning with Regularization}
\label{sec:regularization}

To address the motion style forgetting, we introduce a regularization term that preserves the expert controller's motion style, which is essential for obtaining high-performance bipedal locomotion policies in the real-world.

In the continual learning community~\cite{wang2024acomprehensive}, various approaches have been developed to address catastrophic forgetting, where a pre-trained network loses performance on previous tasks during fine-tuning. These approaches include regularizing either the network weights or the function outputs.
In our case, we choose to regularize the function outputs using MBC similar to~\cite{Ilija2024realworld}:
\begin{equation} 
    \label{eq:reg_ppo}
    L_{FullReg}(\theta, \sigma) =  
    \added{L_{PPO}(\theta, \sigma) + w \mathbb{E_{(\mathbf{s}, \mathbf{a^E}) \sim \mathcal{D}}}\left[\| \mathbf{a}^E - \mu_{\theta}(\mathbf{s})\|_2^2\right]},
\end{equation}
where $\mathbf{a}^E$ is the expert action of the given state $\mathbf{s}$, \replaced{$L_{PPO}$}{$\ell_{PPO}$} is the PPO loss, and $w$ is a weight for the regularization loss. The policy is modeled as Gaussian distribution with the mean $\mu_{\theta}(\mathbf{s})$ and learnable standard deviation $\sigma$.

However, this straightforward regularization can result in a suboptimal policy because the regularization term limits the policy improvement that could be achieved by RL. In the extreme case of \replaced{$w \to \infty$}{$\lambda \to \infty$}, the learning framework simply replicates the expert's behaviors without any improvements. Ideally, the weights should be dynamically adjusted, increasing the importance of the regularization term when the expert shows stable performance and vice versa.

\vspace{-0.5em}
\subsection{Model-Assumption-based Regularization}
\label{sec:assumption-based}

\begin{figure}
\centering
\vspace{1em}
\includegraphics[width=\columnwidth]{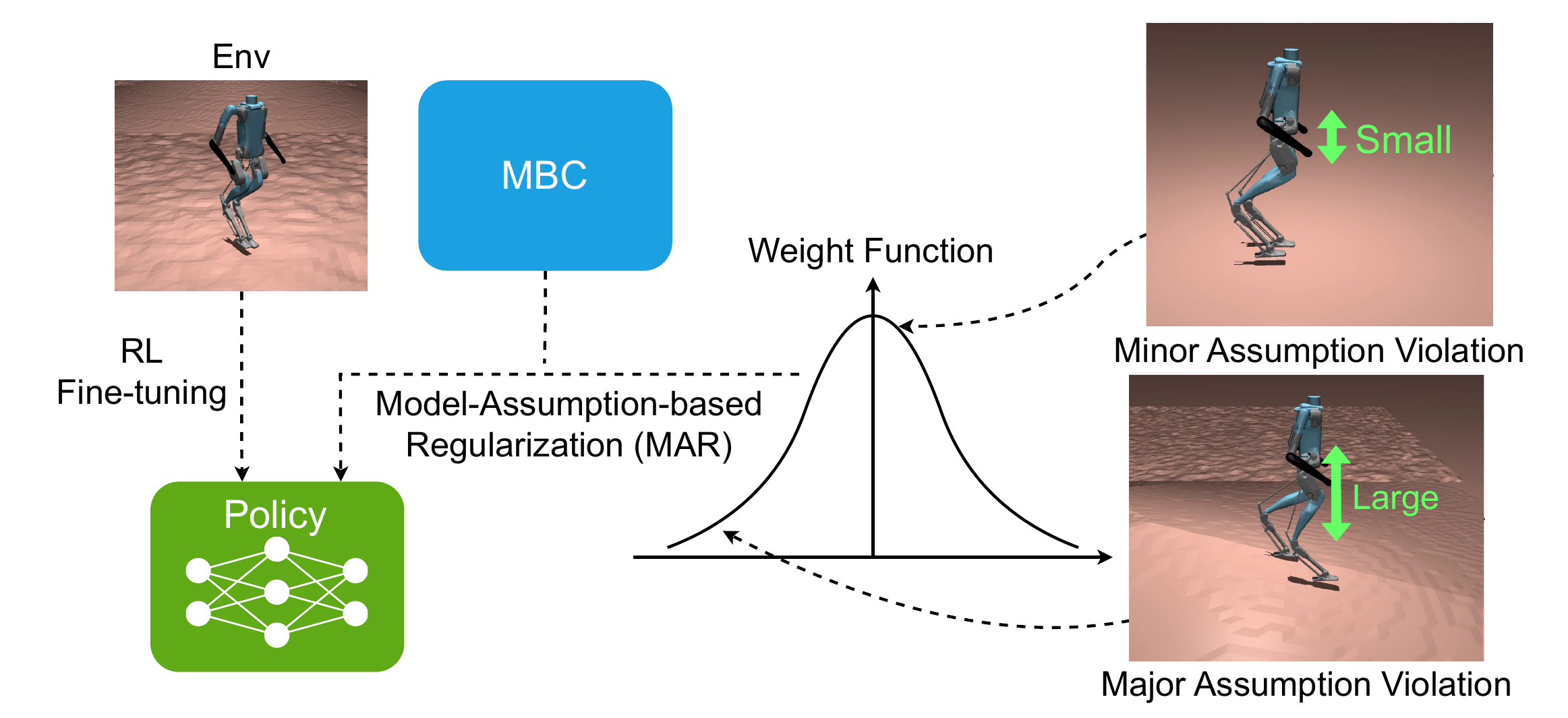}
\vspace{-2.5em}
\caption{Overview of Model-Assumption-based Regularization (MAR). Our framework automatically adjusts the supervised loss regularization based on the assumption violation of MBC.}
\vspace{-2em}
\label{fig:mar}
\end{figure}

To this end, we introduce model-assumption-based regularization (MAR). This approach adjusts the weight of the regularization term by measuring how much the current state violates the assumption of the model-based controller as shown in Fig.~\ref{fig:mar}. Because our model follows the ALIP model assumption~\cite{gong2022zerodynamicspendulummodels} (Eq.~(18)) and incorporates the base height as feedback to the model-based controller, we use the vertical velocity of the base as a criterion to determine whether the current state violates the model assumption.
Then the loss function given by Eq.~(\ref{eq:reg_ppo}) is modified as follows:
\begin{equation}
    \label{eq:preci}
     \added{L_{PPF}(\theta, \sigma) = L_{PPO}(\theta, \sigma) + \mathbb{E_{(\mathbf{s}, \mathbf{a^E}) \sim \mathcal{D}}}\left[w(\mathbf{s}) \| \mathbf{a}^E - \mu_{\theta}(\mathbf{s})\|_2^2 \right],}
\end{equation}
where $w: \mathbf{s} \rightarrow \left[ 0, \infty\right)$ is the weighting function given as follows with the smoothing parameter \added{$\delta$} and coefficient \added{$w_0$}:
\begin{equation}
\label{eq:smooth_reg}
w(\mathbf{s}) = \added{w_0} e^{\frac{-{\dot{z}\added{(\mathbf{s})}}^2}{\added{\delta}}},
\end{equation}
\added{In our case, $w_0=5$ and $\delta=0.0159$ to preserve the motion in reliable states. }This particular choice of the weighting function is based on the observation that the model assumption can be violated even when the controller performs successfully. By smoothly transitioning the weight to zero as violations increase, it dynamically adjusts the importance of expert knowledge for each sample by evaluating it against the given assumption. As a result, when a state deviates from the desirable state due to various factors, such as terrain changes or high command velocities, our framework automatically adjusts the weight of the regularization term for the corresponding states and achieves better performance. Moreover, our approach provides a more direct and intuitive adjustment of the loss function compared to the indirect tuning of reward functions~\cite{Ilija2024realworld, lee2024integrating}. 

\subsection{Implementation Details}
\label{sec:details}
\added{\noindent\textbf{Observation and action space.} The observation space includes root linear and angular velocities, projected gravity, user command, lower-body joint positions and velocities, gait phase and domain, previous action, and the history of lower-body joint positions and velocities. The action space consists of target joint positions and velocities for the lower body.}

\begin{table}[t]
    \centering
    \vspace{0.5em}
    \renewcommand{\arraystretch}{1.0}
    \caption{Reward}
    \label{tab:reward}
    \vspace{-1em}
    \begin{tabular}{l l l}
        \hline
        \textbf{Term} & \textbf{Equation} & \textbf{Weight} \\
        \hline
        Lin Vel Track & $\exp(-\|\mathbf{v}_{xy} - \mathbf{v}_{xy}^{cmd}\|_2^2 / \added{\delta_{xy}})$ & $1.2$ \\
        Ang Vel Track & $\exp(-(\omega_z - \omega_z^{cmd})^2 / \added{\delta_{\omega}})$ & $1.1$ \\
        \hline
        Torque Penalty & $\|\mathbf{\tau}\|_2^2$                                        & $-4\times 10^{-6}$ \\
        Base Motion & $\|\omega_{xy}\|_2^2$ & $-0.6$ \\
        \hline
    \end{tabular}
    \vspace{-2.5em}    
\end{table}
\noindent\textbf{Reward functions.} We design a simple reward function that aims to track the target linear and angular velocities. It also regularizes excessive movements by penalizing torques and base motion. The details are listed in Table~\ref{tab:reward}.

\added{\noindent\textbf{Training Time and Data Size} The model is trained using a GeForce RTX 4090 GPU and an Intel i9-14900K CPU. We use MuJoCo~\cite{todorov2012mujoco} for the simulator. The DAgger stage runs for 380 iterations, corresponding to 9.12 million samples, and takes approximately 5 hours and 26 minutes. The fine-tuning stage completes 10,000 iterations, generating approximately 400 million samples over 31 hours and 33 minutes.}

\noindent\textbf{Lipschitz Continuity Penalty.}
To suppress the vibration of motors, we introduce a Lipschitz continuity penalty as introduced in \citeet{Chen}{chen2024lcp}.
Unlike \cite{chen2024lcp}, we directly penalize the norm of the change in network $\mu_{\theta}$, which is deployed in the testing time:
\begin{equation}
\begin{aligned}
    &\min_{\theta, \sigma} \quad \added{L_{PPF}}(\theta, \sigma) \\
    &\text{s.t.} \quad \max_{\mathbf{s}} \left\| \nabla_{\mathbf{s}} \mu_{\theta}(\mathbf{s}) \right\|^2 \leq \added{K^2},
\end{aligned}
\end{equation}
where \added{$K$} is the Lipschitz constant.
Following the simplification process in~\cite{chen2024lcp}, it is formulated as follows:
\begin{equation}
\begin{aligned}
    \min_{\theta, \sigma} \quad \added{L_{PPF}}(\theta, \sigma) + \alpha \mathbb{E} \left[ \left\| \nabla_\mathbf{s} \mu_{\theta}(\mathbf{s}) \right\|^2 \right]
\end{aligned}
\end{equation}
where $\alpha$ is a tunable variable\added{ set to $\alpha=0.0001$ in our case}. 
\section{Experimental Results}
We design simulation tests and hardware experiments to investigate the following questions: (1) Can PPF learn an effective policy in the training environments? (2) Can PPF show robust performance in sim-to-sim and sim-to-real transfer scenarios compared to the baseline methods?
(3) Can MAR dynamically adjust the sample weights based on the model-based assumption violation?

\subsection{Experimental Details}
\subsubsection{Baselines}
We consider the following baselines for our simulation tests and hardware experiments. All neural network-based controllers use a multi-layer perceptron with hidden layers of sizes 512, 256, and 64.
\begin{itemize}
    \item \textbf{MBC:}  As a model-based control (MBC) baseline, we adopt the passivity-based whole-body controller proposed in~\cite{Shamsah2023Integrated}. In MuJoCo~\cite{todorov2012mujoco}, this controller is adjusted to run at 200 Hz with action space conversion applied, as described in~\cite{youm2023imitating}.
    \item \textbf{IFM:} IFM~\cite{youm2023imitating} learns a policy by imitating the given MBC and fine-tuning it. We employ the same reward configuration and Lipschitz penalty.
    \item \textbf{FullReg:} FullReg fully regularizes the learning with the MBC-labeled actions (Eq.~\ref{eq:reg_ppo}) without MAR.
    \item \textbf{PureRL:} PureRL is trained with reinforcement learning without any pre-training or regularization. Its reward configuration is similar to~\cite{Ilija2024realworld}.
    However, we leverage a swing foot trajectory from an ALIP foot placement controller, add rewards for foot air time and second-order action smoothing, and exclude the selected joint position penalty and target joint position smoothing.

\end{itemize}
\subsubsection{Humanoid}
As a robotic platform, we employ the Digit provided by Agility Robotics. This has the same hardware configuration as described in~\cite{Ilija2024realworld}.
\subsubsection{Simulation}
\label{sec:mujoco_sim_test}
We used MuJoCo~\cite{todorov2012mujoco} for training and testing (Sec.~\ref{sec:mujoco_test}), and the company-provided simulator (AR-Sim) for testing (Sec.~\ref{sec:ar_sim_test}). MuJoCo is a high-performance physics engine designed to simulate complex dynamic systems. AR-Sim offers dynamics that closely mirror the dynamics of the physical robot. Once the training is done in MuJoCo, we deploy policies to either AR-Sim or hardware without additional fine-tuning.

\subsection{Training Environment Experiments}
\label{sec:mujoco_test}

\begin{figure}
\centering
\vspace{0.5em}
\includegraphics[width=0.7\columnwidth]{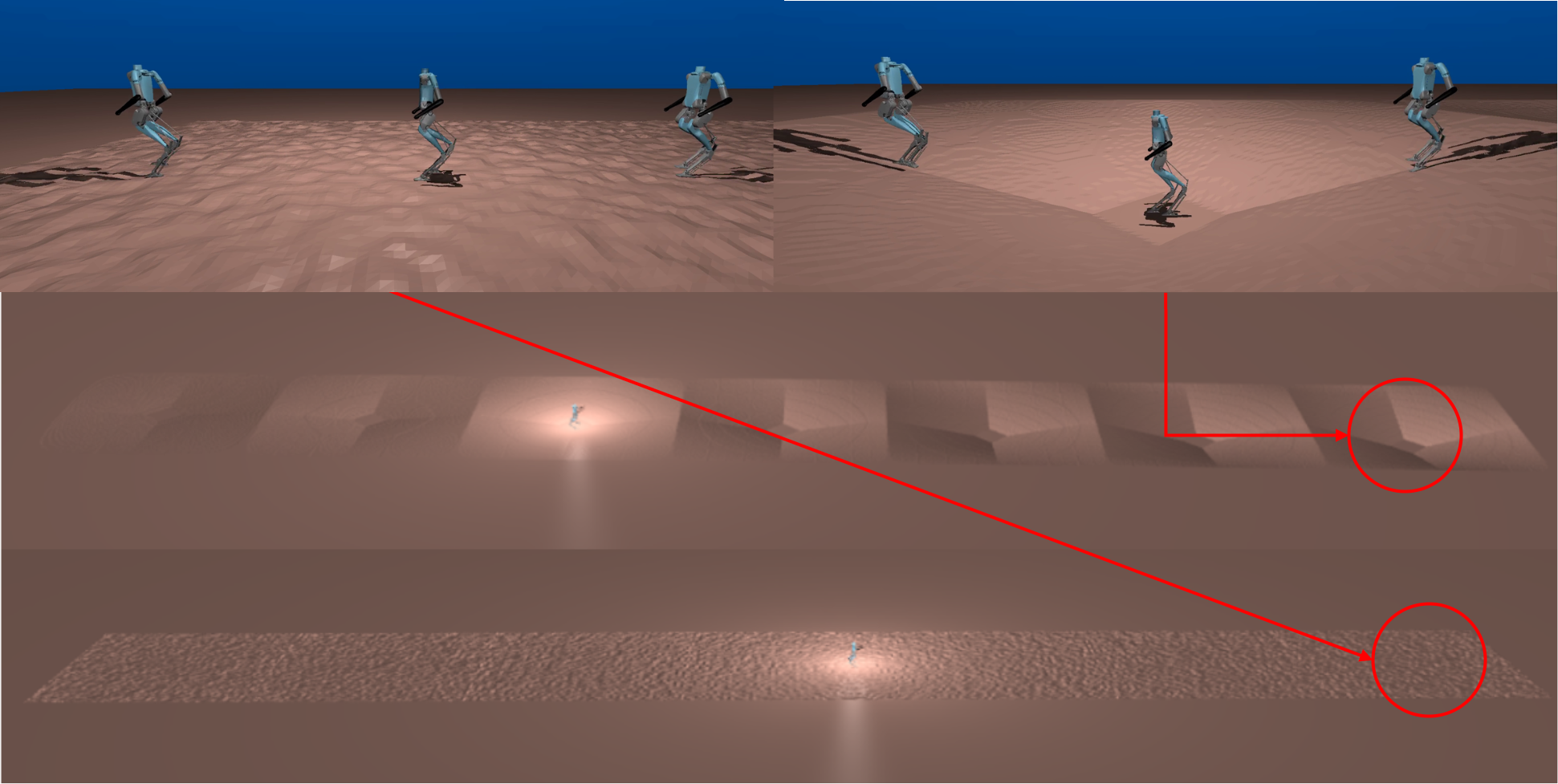}
\vspace{-8pt}
\caption{Testing terrains for the robustness tests in MuJoCo.}
\label{fig:mujoco_test_terrain}
\vspace{-2.2em}
\end{figure}

\begin{figure}
\centering
\vspace{0.5em}
\includegraphics[width=0.9\columnwidth]{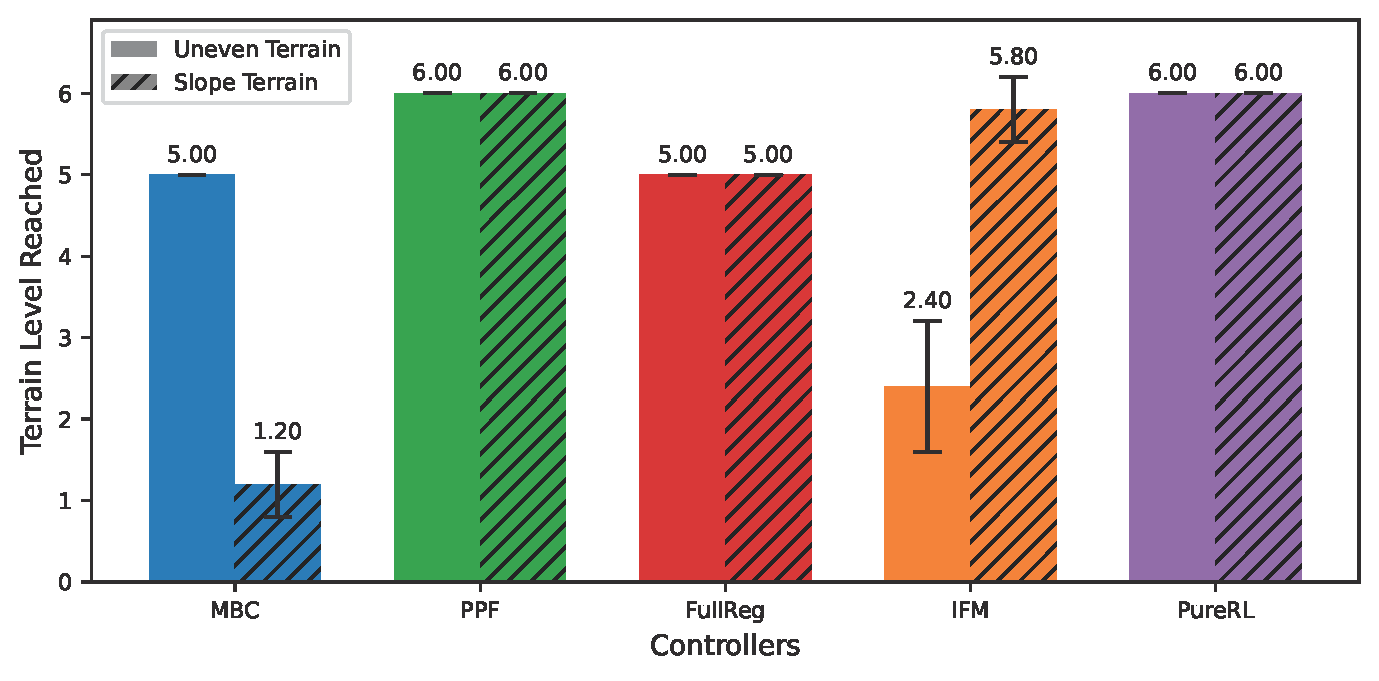}
\vspace{-1.5em}
\caption{Terrain level reached by each controller in the MuJoCo robustness test. PPF successfully traverses all terrains within the time limit. IFM stumbles over its own feet on uneven terrain. FullReg fails to complete the final terrain due to poor tracking performance.}
\label{fig:mujoco_robustness_test_result}
\vspace{-0.5em}
\end{figure}
We evaluate the robustness of the controllers in MuJoCo on sloped and uneven terrains (Fig.~\ref{fig:mujoco_test_terrain}).
The robot is commanded to move forward at 0.6 m/s while adjusting its orientation to maintain a forward direction. As the robot moves forward, it traverses a sequence of terrains with increasing difficulty. The test is conducted with five different random seeds, and each controller runs until the robot either falls, reaches the time limit of 130 seconds, or reaches the end of the last level of the terrain (goal location).

We measure the terrain level reached by each controller shown in Fig.~\ref{fig:mujoco_robustness_test_result}. Overall, PPF (ours) and PureRL show the best performance among all baselines. 
While IFM reaches near the final level on sloped terrain, it often fails on uneven terrain due to unexpected foot contacts. These contacts cause severe lateral and angular disturbances, eventually leading the robot to trip over its own foot and fall.
In contrast, PPF robustly handles such disturbances as a result of its pre-trained motion and consistently reaches the final terrain level in both terrains.
FullReg also demonstrates this level of robustness compared to IFM, but its tracking performance degrades under model-assumption violations, causing it to time out before reaching the goal. A more detailed discussion of this is provided in Section~\ref{sec:effect_of_mar}.
PureRL also reaches the goal within the time limit on both terrains. However, the policy is overfitted to the simulation environment, resulting in degraded motion quality, which may lead to poor performance in both sim-to-sim and sim-to-real transfer. We discuss this further in Sections~\ref{sec:ar_sim_test} and~\ref{sec:hw_exp}.

\subsection{Sim-to-sim Transfer Experiments}
\label{sec:ar_sim_test}
In this subsection, we evaluate sim-to-sim transfer performance through experiments in AR-Sim.

\subsubsection{Robustness Test} \label{sec:ar_sim_robustness}
\newcommand{\rowspace}{\\[-6pt]}
\renewcommand{\arraystretch}{1.4}

\begin{table}[]
\centering
\caption{AR-Sim Robustness Performance}
\label{tab:ar_sim_slope_terrain}
\vspace{-1em}
\resizebox{\columnwidth}{!}{
\begin{tabular}{|c|cccccc|}
\hline
\multirow{3}{*}{\textbf{Method}}
& \multicolumn{2}{c|}{\textbf{10° Uphill}}  
& \multicolumn{2}{c|}{\textbf{12° Uphill}}  
& \multicolumn{2}{c|}{\textbf{14° Uphill}}  
\\ \cline{2-7} 

& \multicolumn{1}{c|}{Succ.}  
& \multicolumn{1}{c|}{Track Err. (\(\downarrow\))}
& \multicolumn{1}{c|}{Succ.}
& \multicolumn{1}{c|}{Track Err. (\(\downarrow\))}
& \multicolumn{1}{c|}{Succ.}
& \multicolumn{1}{c|}{Track Err. (\(\downarrow\))}
\rowspace
& \multicolumn{1}{c|}{(\%)}  
& \multicolumn{1}{c|}{(\%)}
& \multicolumn{1}{c|}{(\%)}
& \multicolumn{1}{c|}{(\%)}
& \multicolumn{1}{c|}{(\%)}
& \multicolumn{1}{c|}{(\%)}
\\ \hline

\textbf{MBC}
& \multicolumn{1}{c|}{0}
& \multicolumn{1}{c|}{\(34.5\)}
& \multicolumn{1}{c|}{0}
& \multicolumn{1}{c|}{\(33.2\)}
& \multicolumn{1}{c|}{0}
& \multicolumn{1}{c|}{\(34.4\)}
\\ \hline
\textbf{\replaced{PPF}{PreCi}}
& \multicolumn{1}{c|}{\textbf{100}}
& \multicolumn{1}{c|}{\(\mathbf{8.4}\)}
& \multicolumn{1}{c|}{\textbf{100}}
& \multicolumn{1}{c|}{\(\mathbf{9.1}\)}
& \multicolumn{1}{c|}{\textbf{100}}
& \multicolumn{1}{c|}{\(\mathbf{18.9}\)}
\\ \hline
\textbf{FullReg}
& \multicolumn{1}{c|}{\textbf{100}}
& \multicolumn{1}{c|}{\(12.1\)}
& \multicolumn{1}{c|}{\textbf{100}}
& \multicolumn{1}{c|}{\(13.1\)}
& \multicolumn{1}{c|}{\textbf{100}}
& \multicolumn{1}{c|}{\(28.7\)}
\\ \hline
\textbf{IFM}
& \multicolumn{1}{c|}{\textbf{100}}
& \multicolumn{1}{c|}{\(15.5\)}
& \multicolumn{1}{c|}{\textbf{100}}
& \multicolumn{1}{c|}{\(24.9\)}
& \multicolumn{1}{c|}{\textbf{100}}
& \multicolumn{1}{c|}{\(30.7\)}
\\ \hline
\textbf{PureRL}
& \multicolumn{1}{c|}{85}
& \multicolumn{1}{c|}{\(10.9\)}
& \multicolumn{1}{c|}{60}
& \multicolumn{1}{c|}{\(15.4\)}
& \multicolumn{1}{c|}{0}
& \multicolumn{1}{c|}{\(54.1\)}
\\ \hline

\end{tabular}
}
\vspace{-2.5em}
\end{table}
We first evaluate robustness by commanding the robot to traverse uphill platforms with three different slopes and measure the success rate and linear velocity tracking error with respect to the commanded velocity, as shown in Table~\ref{tab:ar_sim_slope_terrain}. A trajectory is considered successful if the robot traverses the 5-meter uphill section without falling.

Overall, PPF, IFM, and FullReg demonstrate strong robustness, while MBC and PureRL exhibit significantly degraded performance.
MBC consistently performs the worst, whereas controllers that imitate it (PPF, FullReg, and IFM) show substantially improved robustness. This improvement can be attributed to training the policies on diverse terrains with reinforcement learning objectives, which improves their ability to handle challenging environments. 
Although PureRL successfully navigates the final slope terrain in the MuJoCo test, it achieves much lower success rates in AR-Sim. This highlights the importance of incorporating the periodic gait style of MBC, which facilitates transfer to different simulators and potentially to real-world environments.

Among the methods achieving a 100\% success rate, PPF outperforms the others in tracking accuracy, exhibiting the smallest increase in tracking error as the slope angle increases. In contrast, FullReg shows worse tracking performance, suggesting that full regularization with MBC degrades tracking accuracy in the presence of model assumption errors. We further analyze the effect of MAR in Section~\ref{sec:effect_of_mar} in the presence of model-assumption violation. 

\subsubsection{Random Velocity Tracking Test}
\label{sec:random_vel_track}
\begin{figure}
\vspace{0.5em}
\centering
\includegraphics[width=0.9\columnwidth]{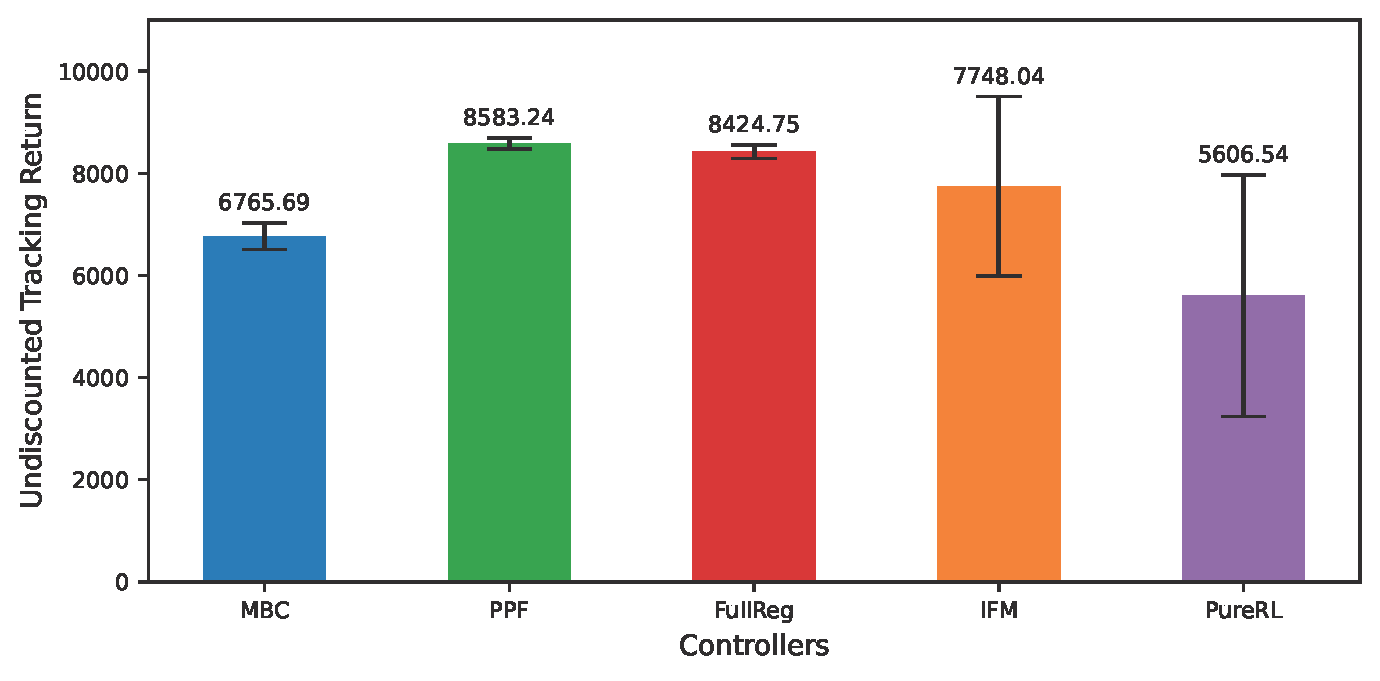}
\vspace{-1.4em}
\caption{Mean undiscounted return of tracking rewards in the AR-Sim random velocity tracking test. The command range is wider than that used during training. Policies learned with regularization, PPF and FullReg, show superior performance compared to those without MBC regularization.}
\label{fig:ar_sim_random_vel_track}
\vspace{-1.6em}
\end{figure}

We further evaluate the tracking performance of each controller by providing four random velocity commands on flat terrain. To assess generalization in the command space, the commands are sampled from a range that includes previously unseen lateral and angular velocities. We measure the mean undiscounted return of linear and angular velocity tracking rewards over 100 trajectories, as shown in Fig.~\ref{fig:ar_sim_random_vel_track}.

PPF outperforms all baselines, achieving a slightly higher return than FullReg. In contrast, IFM shows worse tracking performance than the other regularization-based methods. This is mainly due to the robot frequently tripping over its own foot and falling when given sudden changes in lateral or angular commands.

Despite its robust performance in the training environment, PureRL demonstrates the worst performance among the baselines, performing even worse than MBC. This result aligns with the observations from the previous experiments, supporting the claim that motion distillation from MBC, whether through pre-training alone (IFM) or in combination with regularization (PPF and FullReg), enhances the controller's robustness to domain shifts.

\subsection{Hardware Experiments}
\label{sec:hw_exp}
\newcommand{\cmark}{\ding{51}}%
\newcommand{\xmark}{\ding{55}}%
\renewcommand{\arraystretch}{1.4}
\begin{table}[t]
    \centering
    \vspace{0.5em}
    \caption{Hardware Experiments}
    \vspace{-1em}
    \label{tab:hw_experiment}
\resizebox{\columnwidth}{!}{
    \begin{tabular}{|c|c|c|c|c|c|}
    \hline
    \multirow{3}{*}{\textbf{Method}}& \multicolumn{2}{c|}{Indoor Experiment} & \multicolumn{3}{c|}{Outdoor Experiment} \\
    \cline{2-6} 
    & Max. Forward & Step \& Slip & Slope & Uneven& Sand
    \rowspace
    & Vel. (m/s) & Succ. (\%)& Succ. & Succ. & Succ.\\
    \hline
    MBC     & 0.5 m/s   & 0\%   & \xmark  & \xmark  & N/A \\
    \hline
    \replaced{PPF}{PreCi}   & \textbf{1.5 m/s} & \textbf{100\%} & \textbf{\cmark}  & \textbf{\cmark}  & \textbf{\cmark} \\
    \hline
    FullReg & 1.1 m/s   & 0\%   & \textbf{\cmark}  & \textbf{\cmark}  & \textbf{\cmark} \\
    \hline
    IFM     & $>$1.0 m/s (unsafe) & \replaced{40\%}{67\%}  & N/A  & N/A  & N/A \\
    \hline
    PureRL  & N/A       & 0\%   & N/A  & N/A  & N/A \\
    \hline
    \end{tabular}
    }
    \vspace{-2.5em}
\end{table}

We deploy the controllers on hardware across six different scenarios, both indoor and outdoor, as illustrated in Fig.~\ref{fig:hw_testbed}. The results are summarized in Table~\ref{tab:hw_experiment}.
\subsubsection{Indoor Experiments} In indoor experiments, we measure the maximum forward velocity on flat terrain and the success rate of traversing the step-and-slip platform. This platform consists of a whiteboard covered with poppy seeds or olive oil to induce slipping, and the robot is commanded to traverse it at 0.3 m/s. The success rate is measured over five trials.

PPF achieves the best performance among all baselines, reaching a maximum velocity of 1.5 m/s while maintaining a 100\% success rate in the step-and-slip test. In contrast, MBC and FullReg reach maximum velocities of only 0.5 m/s and 1.1 m/s, respectively. Both controllers fail the step-and-slip test in all trials, slipping in the middle of the whiteboard. Despite being trained with the same RL objective as PPF, FullReg shows limited performance and exhibits a failure pattern similar to MBC.

One possible reason for PPF's performance gain is its ability to handle model assumption violations at higher velocities, whereas MBC requires further tuning of its parameters~\cite{gong2022zerodynamicspendulummodels}. Additionally, by adjusting reliance on MBC action regularization in certain states and placing greater emphasis on the RL objective, PPF better adapts to prevent slips.

IFM struggles with stability during the maximum forward velocity test, frequently tripping over its own foot. For safety, we halted the experiment at 1.0 m/s. PureRL performs the worst in hardware tests. Its maximum velocity could not be measured due to poor tracking performance, and it fails the step-and-slip test as its foot becomes stuck on the whiteboard surface. PureRL struggles to transfer to new domains beyond its training domain unless the motion is carefully designed through extensive reward engineering or it is trained with additional components for sim-to-real, as demonstrated in~\cite{Ilija2024realworld}.

\subsubsection{Outdoor Experiments} We deploy the robot in real-world outdoor scenarios, including sloped, uneven, and sandy terrains. IFM and PureRL are excluded due to unsafe behavior observed even in the controlled indoor testbed.

Both PPF and FullReg successfully traverse all terrain types, including deformable surfaces such as sand, which were not included during training. In contrast, MBC shows limited robustness, struggling even on mildly challenging terrain. Please refer to the supplemental video for details.

\subsection{Effect of MAR in Training and Testing}
\label{sec:effect_of_mar}
\begin{figure}
\vspace{0.5em}
\centering
\includegraphics[width=\columnwidth]{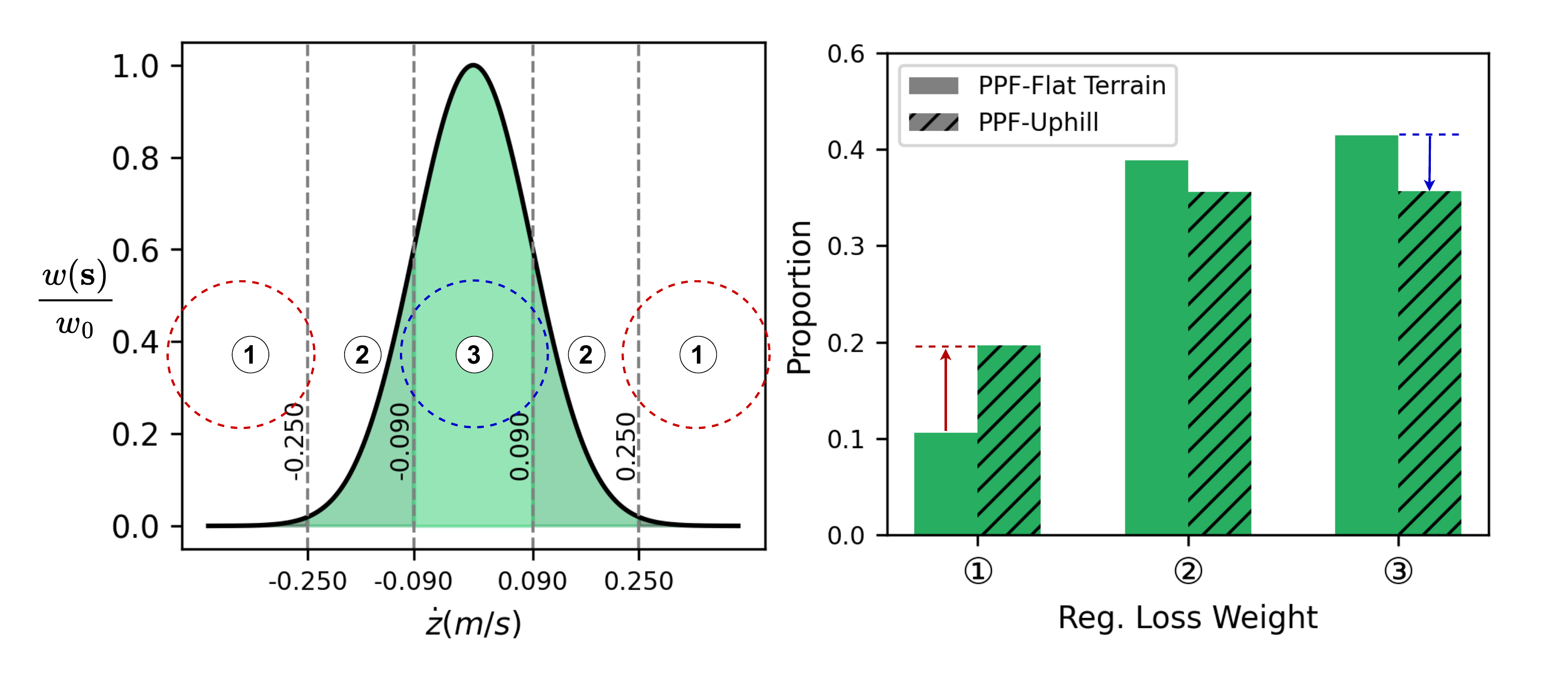}
\vspace{-2.8em}
\caption{\textbf{Left:} an adaptive weight distribution against model assumption violation. \textbf{Right:} a histogram of regularization loss weight during PPF training on both uphill and flat terrain.
We observe more unreliable samples with low regularization weight (region 1) and fewer reliable samples with high weight (region 3) in the uphill case, reflecting how PPF filters out unreliable regularization in violation-prone regions.}
\label{fig:training_z_vel_sup_weight}
\vspace{-1em}
\end{figure}

\begin{figure}
\centering
\includegraphics[width=0.85\columnwidth]{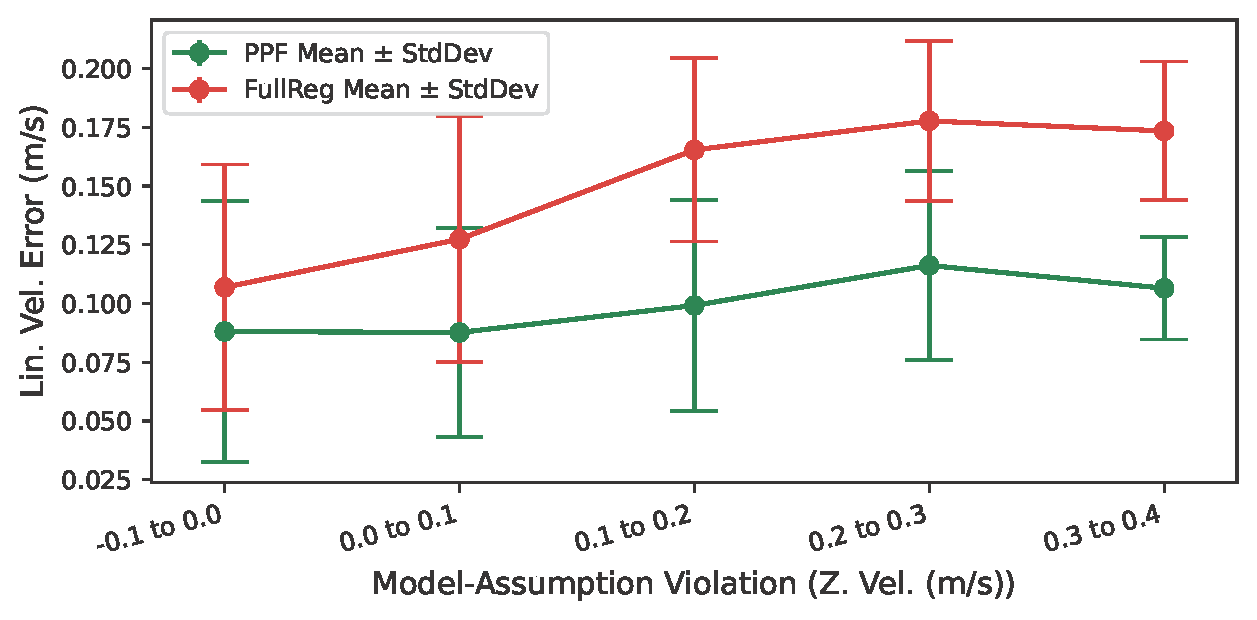}
\vspace{-1.4em}
\caption{Relationship between model-assumption violations and linear velocity tracking errors during a 14-degree uphill test. FullReg exhibits higher tracking errors than PPF when model-assumption violations are more pronounced.}
\label{fig:slope_mar_scatter}
\vspace{-2.1em}
\end{figure}

Although FullReg demonstrates robust performance in both simulation and hardware, its capabilities are often limited compared to PPF when model-assumption violations occur, including the maximum velocity it can reach and tracking ability. We analyze how MAR takes effect when there are model-assumption violations in training and testing. 
\subsubsection{Analysis of MAR at Training}
During training, MAR dynamically reduces the weight of the regularization loss as the model-assumption violation increases, enabling higher performance while preserving the pre-trained motion. In Fig.~\ref{fig:training_z_vel_sup_weight}, we plot the normalized histogram of regularization loss weights across three levels of model-assumption violation, evaluated on both flat and uphill terrains. This illustrates how the weighting adapts dynamically to different scenarios.

Overall, MAR successfully detects model-assumption violations and leverages them to adjust the regularization weights dynamically. On flat terrain, violations are infrequent except under high-velocity commands. In this case, a large proportion of MBC actions remain reliable (region 3 in Fig.~\ref{fig:training_z_vel_sup_weight}), resulting in a higher regularization weight. In contrast, more violations are observed on sloped terrain (region 1 in Fig.~\ref{fig:training_z_vel_sup_weight}), leading to lower weights.
This enables PPF to downweight suboptimal regularization, thereby improving performance in both simulated and real-world environments.

\subsubsection{Analysis of MAR at Testing} 
We examine the relationship between model-assumption violations and linear velocity errors for PPF and FullReg to evaluate the effectiveness of the MAR during testing. Fig.~\ref{fig:slope_mar_scatter} shows the relationship between model-assumption violation (Z-velocity) and linear velocity error in the 14-degree uphill test.

During testing, the model trained with MAR demonstrates consistent performance even in the presence of model-assumption violations. For FullReg, linear velocity tracking error increases as violations become more frequent.

Interestingly, MAR improves not only performance on challenging terrains but also on simpler ones. While PPF performs better on uneven terrain and slopes in MuJoCo simulations, it also outperforms other methods in the Maximum Forward Velocity test on flat terrain and the Step-and-Slip test on hardware, as shown in Table~\ref{tab:hw_experiment}. This improvement may result from increased model-assumption violations caused by high-speed or unstable locomotion even in simple scenarios, or it may reflect PPF's ability to achieve more general performance by ignoring irrelevant samples during training.
\section{CONCLUSIONS}
We propose a novel learning framework, \replaced{PPF}{PreCi}, that combines model-based control and learning-based approaches to effectively train a robust humanoid locomotion control policy. \replaced{PPF}{PreCi} integrates several components, such as pre-training through imitation of a model-based controller, fine-tuning via reinforcement learning, and model assumption-based regularization (MAR), to enhance the robustness and adaptability of humanoid robots across diverse and challenging tasks. Our approach addresses key limitations of prior methods, such as the limited performance of model-based controllers and the catastrophic forgetting in IFM~\cite{youm2023imitating}.

Through extensive simulation tests and hardware experiments on the Digit humanoid robot, we demonstrate that \replaced{PPF}{PreCi} outperforms baseline methods in terrain robustness, velocity tracking, and sim-to-real transfer. \replaced{PPF}{PreCi} achieves a maximum forward velocity of 1.5 m/s and can navigate complex terrains, including slippery surfaces, slopes, uneven ground, and deformable sand, with zero-shot deployment. The incorporation of MAR enables the policy to adapt to scenarios where model assumptions are violated, resulting in superior performance compared to fully regularized methods.

One future extension of this work is to apply MAR to a broader range of model-based assumptions. \replaced{For example, MAR can utilize estimated foot slipperiness through stance foot velocity, as most high fidelity models assume zero contact velocity.}{This includes an in-depth investigation into methods for quantifying model-assumption violations in the single rigid body model.} \added{Furthermore, applying MAR to more sophisticated model-based controllers could provide insights into whether incorporating stronger prior knowledge during pre-training enhances training efficiency and final performance.} \replaced{Another possible}{In addition, another natural} extension is to adapt the framework for more dynamic and high-impact motions, such as running or jumping. This could involve refining the model assumptions or regularization techniques to better handle faster dynamics and increased instability, broadening the framework’s applicability to agile robotic behaviors.
\bibliographystyle{bibtex/IEEEtran}
\bibliography{bibtex/IEEEabrv,bibtex/ref}

\addtolength{\textheight}{-12cm}   


\end{document}